\documentclass{article}
\usepackage{spconf,amsmath,graphicx}

\usepackage{multirow}
\usepackage{ctable}
\usepackage{caption}
\usepackage{subcaption}

\title{object and text-guided semantics for CNN-based activity recognition}

\name{\textsuperscript{$\star$}Sungmin Eum \textsuperscript{$\dagger\mathsection$}, \textsuperscript{$\star$}Christopher Reale \textsuperscript{$\dagger$}, Heesung Kwon\textsuperscript{$\dagger$}, Claire Bonial \textsuperscript{$\dagger$}, Clare Voss\textsuperscript{$\dagger$} \thanks{\textsuperscript{$\star$}These authors contributed equally to this work}}
\address{
         \textsuperscript{$\dagger$}U.S. Army Research Laboratory, Adelphi, MD, USA\\
         \textsuperscript{$\mathsection$}Booz Allen Hamilton Inc., McLean, VA, USA\\
         }
\begin{document}
%
\maketitle
\begin{abstract}
Many previous methods have demonstrated the importance of considering semantically relevant objects for carrying out video-based human activity recognition, yet none of the methods have harvested the power of large text corpora to relate the objects and the activities to be transferred into learning a unified deep convolutional neural network. We present a novel activity recognition CNN which co-learns the object recognition task in an end-to-end multitask learning scheme to improve upon the baseline activity recognition performance. We further improve upon the multitask learning approach by exploiting a text-guided semantic space to select the most relevant objects with respect to the target activities. To the best of our knowledge, we are the first to investigate this approach.

\end{abstract}
\section{Introduction}
\label{sec:intro}

In recent years, a significant amount of research in the computer vision community has focused on human activity recognition. The objective of this research is to be able to automatically recognize and understand what humans depicted in a video are doing. In this work, human activity recognition is formulated as a classification problem (i.e., given a short video clip, which activity in a given set is depicted). This problem is important for several applications including human-robot teaming (helping robots to understand and interact with their environment and thus better react to it), surveillance (sift through a large number of video streams to detect abnormal behavior), and video-tagging (automatically tag videos to make them easier to find).

Most state-of-the-art approaches to this problem train deep convolutional neural networks (CNN) to classify videos based on their raw pixels and/or extracted features. One prominent method, the two-stream approach \cite{simonyan2014two}, trains one network to classify single RGB frames and a second network to classify short snippets of optical flow features. Temporal Segment Networks \cite{wang2016temporal} attempt to exploit longer-term temporal information by grouping frames from different portions of the video during training. Karpathy et al. \cite{karpathy2014large} apply deep learning to a very large dataset. In addition to the standard convolutional neural networks, three-dimensional convolutional neural networks \cite{tran2015learning, ji20133d, sun2015human} have been used for activity recognition to great effect.

One of the main limitations of deep learning approaches to this problem is their dependence on the size and scope of the training data set. It is important to have a large labeled training set to take full advantage of the power of deep neural networks. However, it may not be feasible to attain a large enough dataset as it requires excessive human effort to collect and annotate the training videos. Although state-of-the-art methods achieve good results on benchmark datasets, due to limited data, they cannot use the recently proposed deeper network architectures such as ResNet\cite{he2016resnet} without overfitting. Furthermore, regardless of the size of the dataset, it is virtually impossible to guarantee that all variations of the target activities are captured by given video dataset.

In this work, we attempt to address these problems by introducing a novel CNN network which incorporates text-guided object information within a multitask learning scheme. This approach helps the overall network training in two ways. First, it allows us to exploit a large object recognition dataset to boost the amount of training data we have. Second, it allows us to incorporate general knowledge from text about the target activities that may not be fully apparent from the training videos, and our approach results in improvement upon the baseline activity recognition performance.

More specifically, we train our network to simultaneously perform object recognition with the activity recognition. Learning a single model which considers multiple related tasks is known as multitask learning \cite{caruana1997multitask, ranjan2017all, zhang2012robust, yuan2012visual}. In our approach, we are considering object recognition as a highly related task with respect to the activity recognition. Using this approach allows us to leverage the ImageNet dataset \cite{deng2009imagenet}, which provides us with significantly more training data. This enables us to use much deeper networks than current methods in the literature without overfitting and thus achieve higher recognition rates. We further improve upon the multitask learning approach by analyzing the relationship between the activities and the objects within a text-guided semantic space.




\section{Our Method}
\label{sec:method}





\begin{figure}
\begin{center}
   \includegraphics[width=0.95\linewidth]{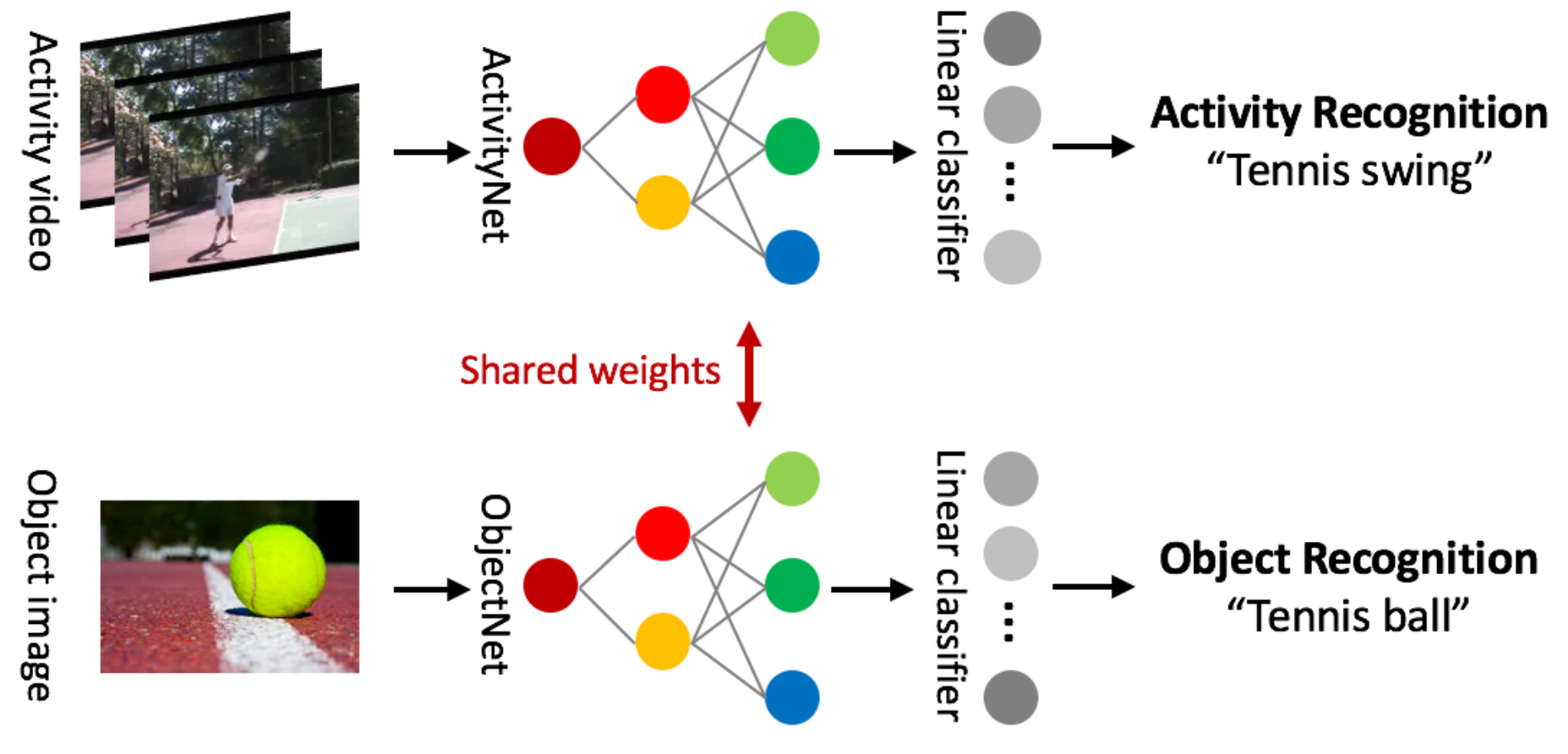}
\end{center}
   \caption{{\bf Object-incorporated activity recognition network architecture.} Colored region of the network is being shared between the ActivityNet and the ObjectNet, while the grayscale portions (softmax linear classifiers) are learned separately to handle the specific tasks of activity and object recognition.}
\label{fig:simpleArchitecture}
\end{figure}

\subsection{Incorporating Object Recognition with Activity Recognition in Multitask Learning}
\label{ss:object-incorporated}

\begin{figure}[htb]
\begin{subfigure}[htb]{\linewidth}
\includegraphics[width=\linewidth]{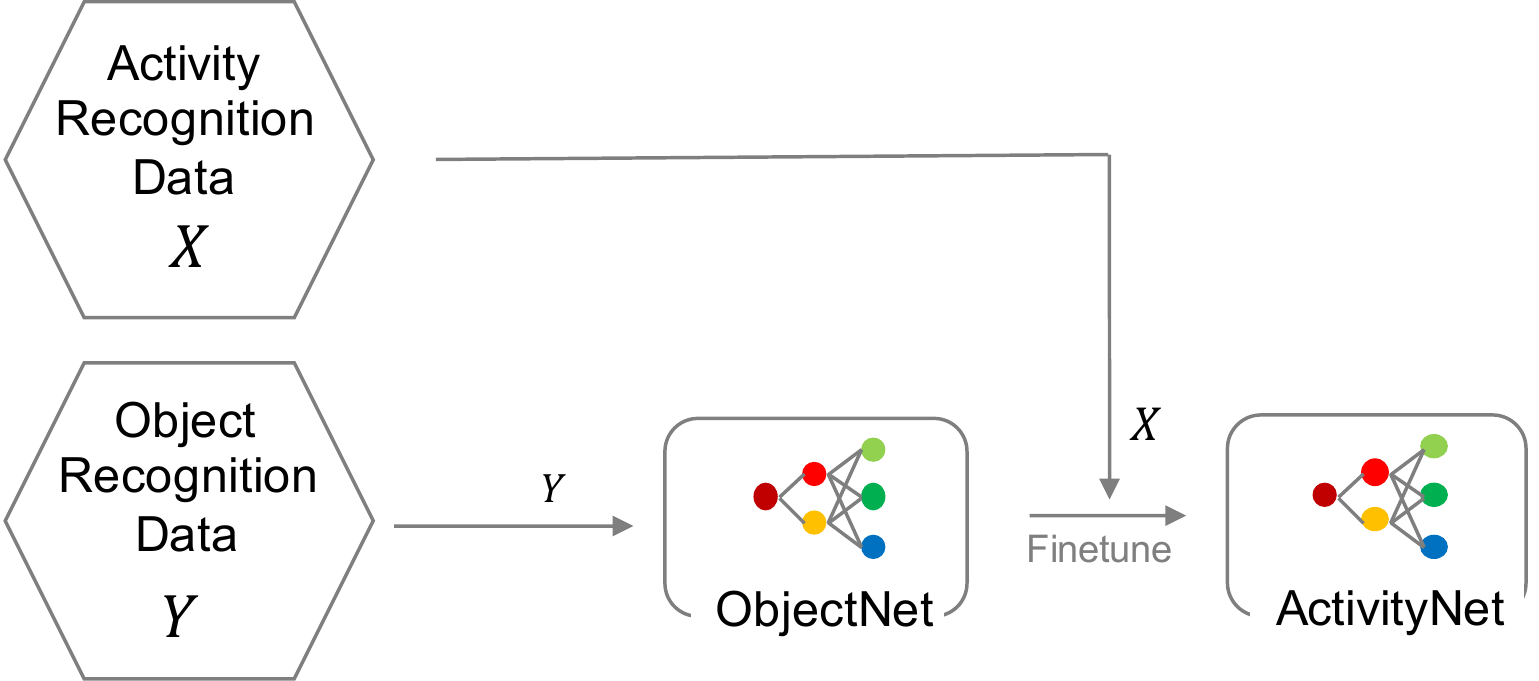}
\subcaption{}
\label{fig:baseline}
\end{subfigure}

\begin{subfigure}[htb]{\linewidth}
\includegraphics[width=\linewidth]{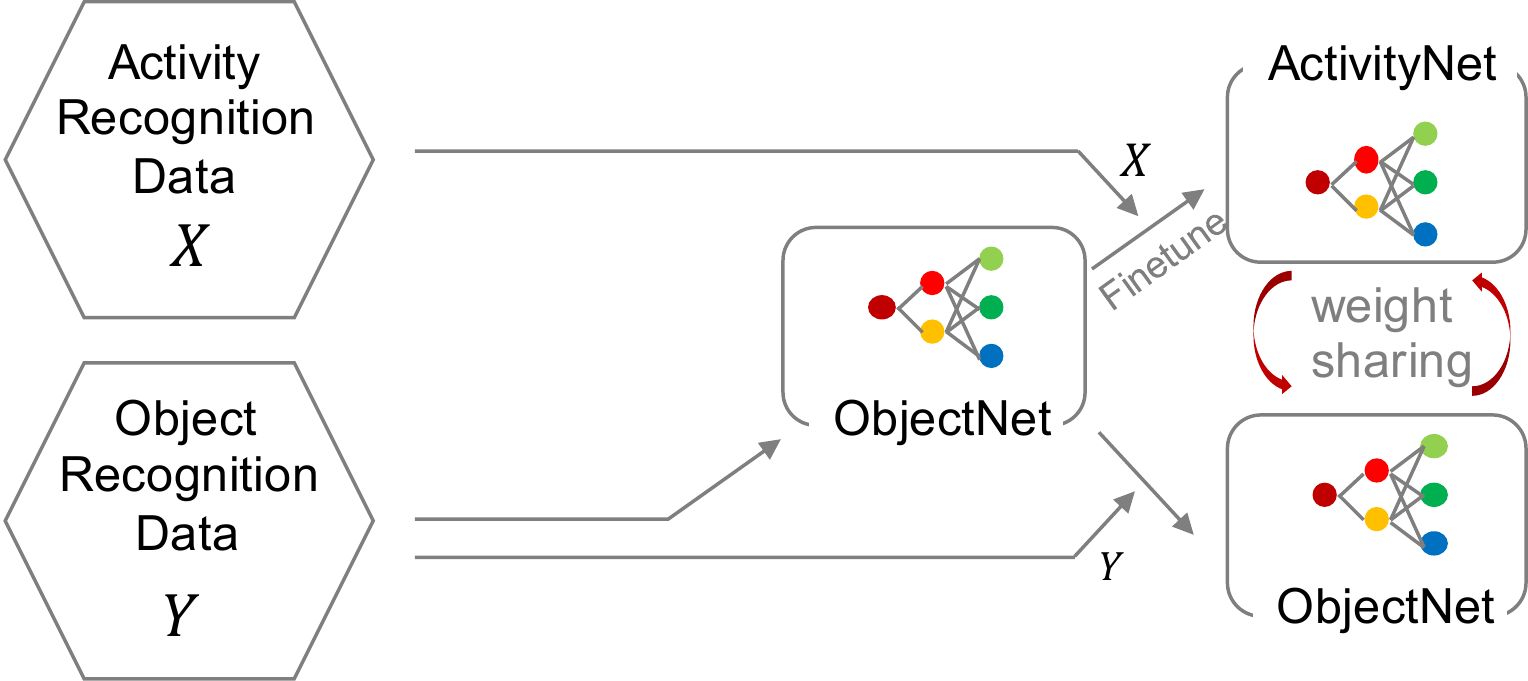}
\subcaption{}
\label{fig:objectincorporated}
\end{subfigure}

\begin{subfigure}[htb]{\linewidth}
\includegraphics[width=\linewidth]{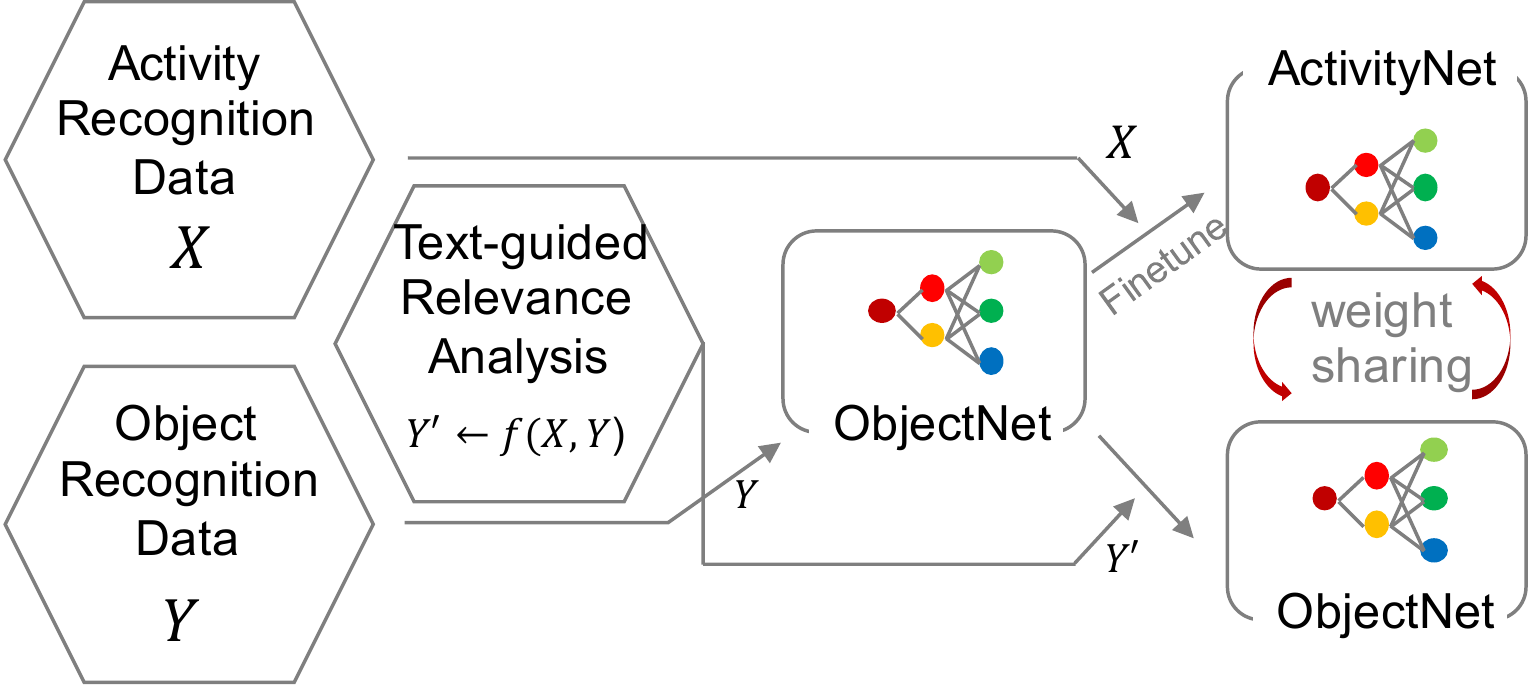}
\subcaption{}
\label{fig:textguided}
\end{subfigure}
\vspace{-0.4cm}
\caption{{\bf Activity recognition network training strategies.} ActivityNet and ObjectNet refer to CNNs for recognizing the activities or objects, respectively. (a) Baseline activity recognition network which finetunes from the pretrained ObjectNet (b) Object-incorporated activity recognition network (c) Text-guided, object-incorporated activity recognition network.} \label{fig:architectures}
\end{figure}

Previous approaches have demonstrated that being able to detect or recognize objects within an image can improve recognition of relevant events and activities in that image \cite{escorcia2013,eum2017iod}. We take a similar approach in exploiting the object information but with two major novel aspects. First, we introduce a practical way of training and enhancing the activity recognition network by carrying out the multitask learning with the object recognition network. Moreover, unlike the previous approaches, we do not attempt to localize or identify the objects within the target domain (in our case, activity recognition) but train the network to perform the task of object recognition using a totally different dataset (ImageNet). This bolsters the amount of training data for the overall network, and at the same time, removes the need for manually annotating/detecting the relevant objects in the target videos. As shown in Figure \ref{fig:simpleArchitecture}, we share the weights in all the layers of the network between the two tasks except the task-specific softmax classifiers. 

Datasets we use for the training (UCF101 \cite{soomro2012ucf101} and ImageNet \cite{deng2009imagenet}) are only annotated for each single task (i.e., videos frames for activity recognition and ImageNet images for object recognition). Thus, we design the network so that each data sample is directly associated with the loss function for the corresponding task. However, as we ground our method in the relevance of the two tasks, all the layers except the softmax layer are being shared between the two tasks.

We can view our multitask learning approach as an extension of the standard finetuning strategy (Figure \ref{fig:baseline}). In training our network we learn the parameter weights for both the activity recognition (ActivityNet) and the object recognition (ObjectNet) by finetuning from the network pretrained for the task of the object recognition (ObjectNet) as shown in Figure \ref{fig:objectincorporated}. The continuation of the incorporation of gradients from the object recognition loss acts as a regularization for the overall network parameters, preventing them from overfitting to the activity recognition task. As our pretrained ObjectNet, we have used the network which was trained to classify 1000 object classes assigned by the ImageNet Challenge \cite{deng2009imagenet}.

\subsection{Leveraging the text-guided semantic space}
\label{sec:textguide}

\begin{figure}
\begin{center}
   \includegraphics[width=\linewidth]{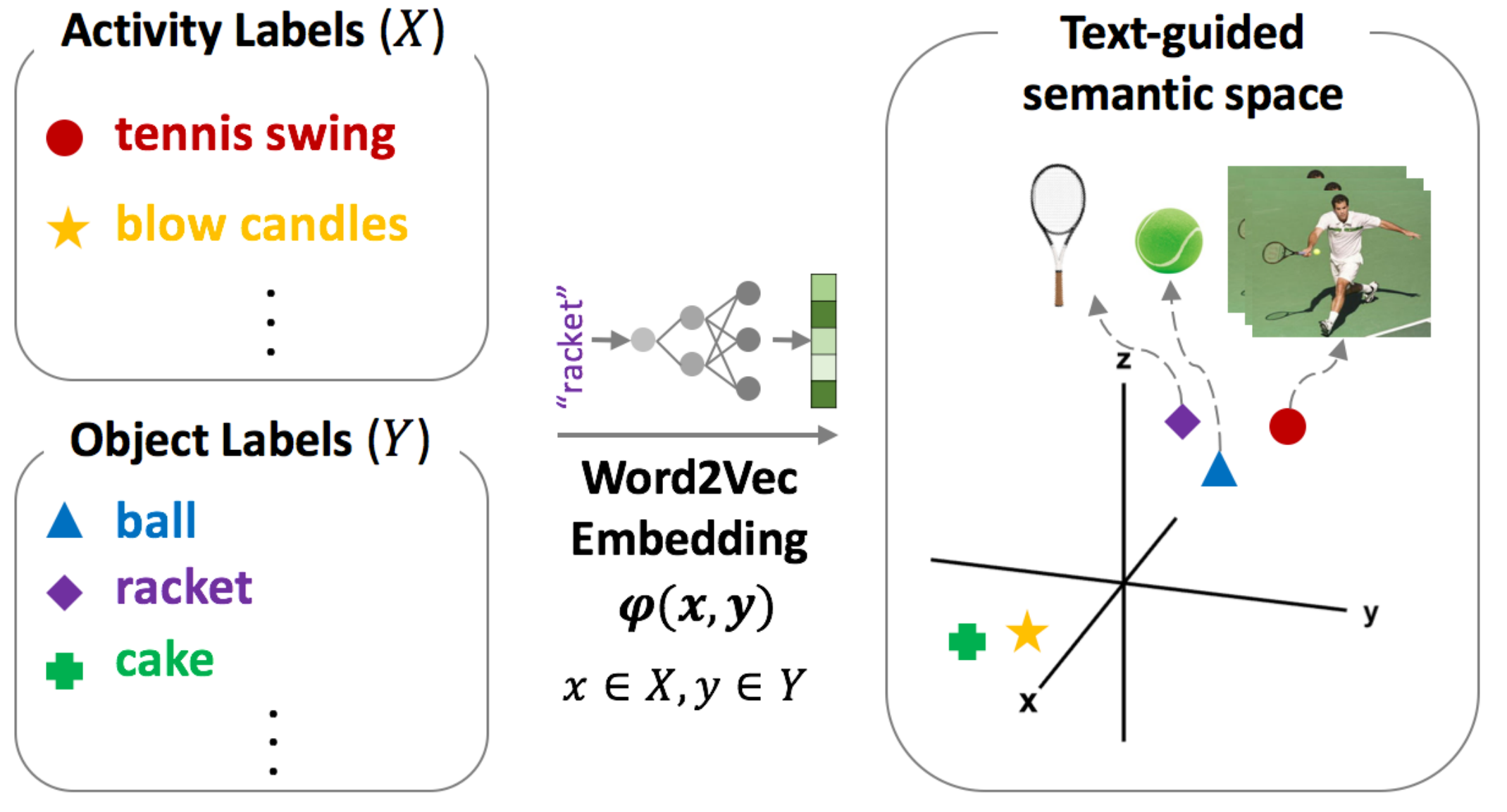}
\end{center}
   \caption{{\bf Text-guided Relevance Analysis in the semantic space.} Closely related activities and objects are aggregated in the text-guided semantic space using the Word2Vec embedding. In our experiments, object labels correspond to ImageNet class labels.}
\label{fig:semanticspace}
\end{figure}

The object-incorporated activity recognition network introduced in Section \ref{ss:object-incorporated} uses all the objects from the ImageNet dataset to learn the ObjectNet, and thus solely relies on the capability of the multitask network learning process to harvest the necessary information about the objects with respect to the activities. We seek to further improve upon our object-incorporated activity recognition network by exploring the following questions: Which objects are more important and indicative for certain activities? Would selecting this subset of objects help improve activity recognition? 

Our strategy is to refine the original object dataset before proceeding into the network training by selecting the most relevant set of objects with respect to the activities in the target domain. To select the most relevant objects, we carry out what we call `Text-guided Relevance Analysis (TRA)' where we compute the similarity between the textual labels of the activities and those of the ImageNet objects within a semantic vector space. We exploit the the textual labels which are originally provided from both datasets ({\it UCF101} and {\it ImageNet}). 

In TRA, we use Word2Vec \cite{mikolov2013distributed} embedding to project the textual labels to the semantic vector space. Word2Vec embeds words and phrases into a vector space based on their usage in a large text corpora. Words that are used in similar contexts will be embedded closer together in the vector space. An illustration of the text-guided semantic space is shown in Figure \ref{fig:semanticspace}, where the activity label ``tennis swing'' is closely embedded with the object labels ``ball'' and ``racket''.

Assuming $\omega(\cdot)$ as the embedding learned by Word2Vec, we approximate the relevance between a target activity $x$ and an ImageNet class $y$ with the cosine similarity of their vector space representations as follows:

\begin{equation}
\varphi(x,y) = \frac{\omega(x)\boldsymbol{\cdot} \omega(y)}{\|\omega(x)\|_2\|\omega(y)\|_2}.
\end{equation}


We then compute the overall relevance $\kappa$ of an ImageNet class $y \in Y$  to the set of target activities $X$ as the sum of the relevances of $y$ to each activity $x \in X$,

\begin{equation}
\kappa(y|X) = \sum_{x \in X} \varphi(x,y).
\end{equation}

\begin{table}[t]
\setlength{\tabcolsep}{7.0pt}
\renewcommand{\arraystretch}{1.4}
\begin{center}
\caption{{\bf Highly ranked ImageNet classes using TRA.} Top 3 ImageNet classes for a set of selected activity classes.}
\label{tab:rankedObjts}
{\small
\begin{tabular}{cccc}
\specialrule{.13em}{.05em}{.05em}
Activity (UCF101) & 1st & 2nd & 3rd  \\ 
\specialrule{.13em}{.05em}{.05em} 
 ApplyLipstick & lipstick & mascara & nail polish \\ \hline
Biking & bicycling & cycling & motorcycle \\ \hline
Knitting & quilting & needlework & knit \\ \hline
MilitaryParade & soldier & Marine & admiral \\ \hline
cliffDiving & cliff & dive & ledge  \\ \specialrule{.13em}{.05em}{.05em} 
\end{tabular}
}
\end{center}
\end{table}

Once we acquire $\kappa(\cdot)$ for all ImageNet classes, we select the most relevant classes (those whose relevance score is numerically highest) to be used for training the ``text-guided, object-incorporated activity recognition network''. This overall process of TRA (See Figure \ref{fig:textguided}), can be considered a dataset refinement procedure $f(\cdot)$ for the original object recognition dataset $Y$ as $Y' = f(X,Y)$:


\begin{equation}
Y' = f(X,Y) = \{ y : rank(\kappa(y|X))  \leq m, y \in Y\},
\end{equation}
where $rank(\kappa(y|X)$) indicates the rank in descending order among all $\kappa(y|X)$ such that $x \in X$ and $y \in Y$, while $m$ is the number of selected objects within $Y$. Based on an empirical analysis, we selected, for our image input dataset (identified as $Y$ in Figure \ref{fig:textguided}), the images that have text-labels for 1000 objects ($m$ = 1000) for training the final version of the network. In Table \ref{tab:rankedObjts}, we introduce some samples of highly ranked object (ImageNet) classes with respect to the activity (UCF101) classes acquired by the TRA.

\section{Experiments}
\label{sec:experiments3}

\subsection{Experimental details}

\noindent{\bf Preprocessing the data.} 
First, we subtract a mean pixel from each pixel in the image. Then we select a random window from the target frame. The window's width and height are randomly and independently selected (from a uniform distribution) to be between 168 and 256 pixels. Once the width and height are selected, the location of the window within the image is selected at random (again, from a uniform distribution). Finally, the window is resized to 224$\times$224 pixels and fed into the network. The random window selection process helps to generate more variation in the training data to reduce the risk of overfitting. For the ImageNet images, we still subtract the pixel mean, but select a sub-image by simply choosing a random 224$\times$224 window from the image. We can use a simpler window selection with ImageNet because it contains many more images which are uncorrelated unlike the video frames which are highly correlated. \\

\noindent{\bf Network architecture setting.}
We use the ResNet \cite{he2016resnet} architectures (ResNet 50, 100, and 152) which has recently demonstrated the state-of-the-art performance in various applications. This is in contrast to previous approaches which use shallower networks. Our multitask approach acts as a regularization, enabling us to use the deeper, better-performing ResNet networks. All networks are initialized by pretraining on the 1000 ImageNet challenge  classes. 

We incorporate the Temporal Segment Network (TSN) \cite{wang2016temporal} approach in training our networks which is known to capture long-term temporal information. We have empirically determined the optimal number of segments to be three, and thus the size of the activity recognition portion of the batch was set to be a multiple of three. For example, when training our ResNet 50 network, total batch size is 64. Ideally, we would split it evenly between the two network streams (32 each). However, as 32 is not a multiple of three, we use 33 activity recognition samples and 31 ImageNet samples.\\


\noindent{\bf Training strategy.}
We train our networks with stochastic gradient descent on single GPU (NVIDIA TITAN X) system. Due to the depths of the networks used and the memory limitations of the GPU (12 GB), we were forced to use small batch sizes of 64, 48, and 32 frames/images for ResNet 50, 101, and 152, respectively. When training in the multitask setting, we split the batch size between activity recognition frames and ImageNet images. We found that splitting the batch approximately evenly between the two (i.e., giving equal weight to the two objectives) provided the best performance.

When training ResNet 50, we initialize the learning rate to .001. We divide it by 10 after 10k and 13k iterations and train for 15k iterations in total. Due to the smaller batch sizes, we initialize the learning rates for the ResNet 101 and 152 to .0005. For ResNet 101, we divide it by 10 after 13k and 18k iterations and train for 20k iterations in total. For ResNet 152, we divide it by 10 after 28k and 36k iterations and train for 40k iterations in total. Weight decay was set as .0001. During training of all three architectures, we place dropout layers just before the final softmax classifiers. Dropout rate is set as .25.

At test time, we use the standard approach of generating predictions for 25 evenly spaced frames. For each frame, we generate predictions from 10 different 224$\times$224 pixel windows: one from each corner of the frame, one from the center of the frame, and then a horizontally flipped version of each of those. For each video, 250 probability predictions are made for each of the classes. We average them and predict the activity with the highest value. In this work, we use a pre-trained Word2Vec model which was trained on an internal Google dataset of news articles containing a billion words \cite{mikolov2013distributed}.

\subsection{Performance Evaluation}

We evaluate the performance of our approach on the UCF 101 benchmark dataset \cite{soomro2012ucf101}. We have used the ResNet to construct the baseline architecture for both the activityNet and the objectNet (See Figure \ref{fig:simpleArchitecture}).
The experiments were carried out on three different ResNet networks (ResNet 50, 101, and 152) under three different settings (baseline, object-incorporated, text-guided + object-incorporated). The baseline approach is the standard method without multitask learning. For the object-incorporated multitask approach, we randomly selected 1000 ImageNet classes to learn the objectNet. The text-guided + object-incorporated approach uses Word2Vec to select the 1000 most relevant ImageNet classes as described in Section
\ref{sec:textguide}.

\begin{table}[t]
\setlength{\tabcolsep}{7.0pt}
\renewcommand{\arraystretch}{1.4}
\begin{center}
\caption{{\bf Performance comparison.} Accuracy on UCF 101 Dataset. See Figure \ref{fig:architectures} for the different training strategies.}
\label{tab:ucf}
{\small
\begin{tabular}{cccc}
\specialrule{.13em}{.05em}{.05em} 
& Baseline & object incorp. & object incorp.\\ & &  & + text-guided \\ 
\hline
 Multitask? & No & Yes & Yes \\
 \specialrule{.13em}{.05em}{.05em} 
ResNet 50 & 81.3 & 84.0 & 85.1\\ \hline
ResNet 101 & 82.6 & 85.3 & 86.9\\ \hline
ResNet 152 & 83.1 & 86.0 & \textbf{87.5} \\ \hline
TSN \cite{wang2016temporal} & 85.7 & - & - \\ \specialrule{.13em}{.05em}{.05em} 
\end{tabular}
}
\end{center}
\end{table}

From the results shown in Table \ref{tab:ucf}, it is clear that using the ResNet networks with the baseline approach provides worse performance than the state-of-the-art method (TSN \cite{wang2016temporal}). This is because the architecture used in \cite{wang2016temporal} uses shallower networks which are not as prone to overfitting. When we incorporate the object information in a multitask learning scheme (object-incorporated), the performance increases close to the current state-of-the-art. And finally, when we exploit the text-guided supervision on top of the object incorporation, we are able to outperform the state-of-the-art.


\section{Conclusion}
\label{sec:Conclusion}

We have introduced a novel way of constructing an object-incorporated and text-guided CNN to better handle the task of video-based human activity recognition. We do this by leveraging the text-guided semantic space to select the most commonly associated objects with respect to the target activities. We then train the network to recognize the target activities as well as the selected set of objects by exploiting a shared network and a multitask learning approach. We have experimentally verified that the strategies of incorporating objects for activity recognition and text-guided object selection are both effective in improving the performance for the human activity recognition. In the future, we are seeking to incorporate the background scenes into our framework as it also carries significant semantic information for the activities.

\bibliographystyle{IEEEbib}
\bibliography{refs}

\end{document}